\documentclass[a4paper,conference]{IEEEtran}
\IEEEoverridecommandlockouts
\usepackage{cite}
\usepackage{amsmath,amssymb,amsfonts}
\usepackage{algorithmic}
\usepackage{graphicx}
\usepackage{textcomp}
\usepackage{xcolor}

\usepackage{subcaption}
\usepackage{arydshln}
\usepackage{hyperref}
\begin{document}

\title{FourierNet: Compact Mask Representation for Instance Segmentation Using Differentiable Shape Decoders\\
\thanks{\IEEEauthorrefmark{1}These authors contributed equally to this work}
} 

\author{\IEEEauthorblockN{Hamd ul Moqeet Riaz\IEEEauthorrefmark{1}, Nuri Benbarka\IEEEauthorrefmark{1} and Andreas Zell}
\IEEEauthorblockA{ Cognitive Systems, 
	\textit{Department of Computer Science (WSI)} \\
\textit{University of T\"ubingen, Germany}\\
Email: hamd.riaz@uni-tuebingen.de, nuri.benbarka@uni-tuebingen.de,\\ andreas.zell@uni-tuebigen.de }
}

\maketitle

\begin{abstract}
We present FourierNet, a single shot, anchor-free, fully convolutional instance segmentation method that predicts a shape vector. Consequently, this shape vector is converted into the masks' contour points using a fast numerical transform. Compared to previous methods, we introduce a new training technique, where we utilize a differentiable shape decoder, which manages the automatic weight balancing of the shape vector's coefficients. We used the Fourier series as a shape encoder because of its coefficient interpretability and fast implementation. FourierNet shows promising results compared to polygon representation methods, achieving 30.6 mAP on the MS COCO 2017 benchmark. At lower image resolutions, it runs at 26.6 FPS with 24.3 mAP. It reaches 23.3 mAP using just eight parameters to represent the mask (note that at least four parameters are needed for bounding box prediction only). Qualitative analysis shows that suppressing a reasonable proportion of higher frequencies of Fourier series, still generates meaningful masks. These results validate our understanding that lower frequency components hold higher information for the segmentation task, and therefore, we can achieve a compressed representation. Code is available at: \href{https://github.com/cogsys-tuebingen/FourierNet}{github.com/cogsys-tuebingen/FourierNet}. 
\end{abstract}


\section{Introduction}
\label{sec:intro}

\quad With the recent emergence of deep learning, combined with readily available data and higher computational power, the use of autonomous machines has become a realistic option in many decision making processes. In applications such as autonomous driving and robot manipulation, the first and foremost task is to perceive and understand the scene before a decision is made.  
\begin{figure}[t]
	\centering
	\includegraphics[width=0.95\linewidth]{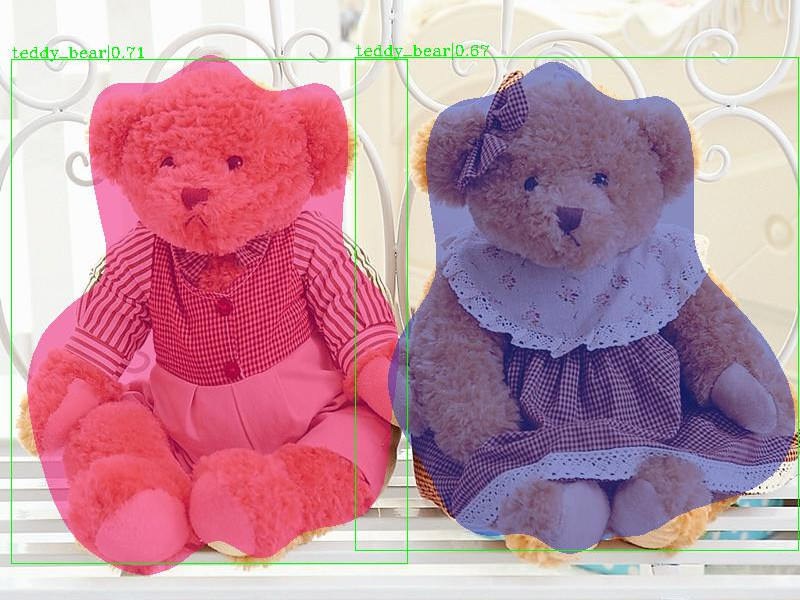}
	\caption{Mask predicted by using \textbf{10 coefficients} (20 parameters) of Fourier series. Note that 90 contour points are used to generate this mask.}
	\label{fig:10coe}
\end{figure}

Instance segmentation is one of the techniques used for scene understanding \cite{janai2017computer}. It categorizes each pixel/region of an image by a specific class and, at the same time, distinguishes different instance occurrences. 
Among the instance segmentation methods are \textit{two-stage} methods that produce a bounding box and then classify the pixels within that box as foreground or background \cite{he2017mask,liu2018path,huang2019mask,chen2019hybrid,lee2019centermask,kuo2019shapemask}. Although these are still dominant in terms of prediction accuracy, they are computationally intensive. 

There is a growing trend to use more straightforward, faster \textit{single-stage} instance segmentation methods that do not require initial bounding box proposals \cite{bolya2019yolact,xie2019polarmask,xu2019explicit,ying2019embedmask,zhou2019bottom}. 
In one of the latest approaches, ESE-Seg \cite{xu2019explicit} have encoded the objects' contours using function approximations such as Chebychev polynomials and Fourier series.
They trained a network to predict a shape vector (a vector of coefficients), in which a numerical transform converts it into contour points in the polar representation. The main advantage of this method is that it requires fewer parameters to represent the mask as opposed to the binary grid or polygon representations \cite{liang2019polytransform}. However, ESE-Seg \cite{xu2019explicit} regresses the shape vector directly. We argue that the direct regression of the shape vector does not weigh each coefficient according to its impact on the mask and prevents the model from learning the actual data distribution.

Therefore, we propose an alternative training method in which the network outputs are passed through a \textit{differentiable shape decoder} to obtain contour points that are used to calculate the loss. In this case, the losses of other polygon representation methods, e.g. polar IOU loss \cite{xie2019polarmask} and Chamfer distance loss \cite{fan2017point}, can be used, and the network is trained for its main task. The gradients of these losses are back-propagated through the decoder and the weight balancing of the different shape vector's coefficients is done automatically.

\section{RELATED WORK}
\label{sec:related}

\subsection{Two stage instance segmentation}
\label{subsec:tsm}
\quad Two-stage instance segmentation splits the task into two subtasks, object detection then segmentation. The most prominent instance segmentation method is Mask R-CNN \cite{he2017mask}, which is constructed on top of Faster R-CNN \cite{ren2015faster} by adding a mask branch parallel to the bounding box and the classification branches. further, they used RoI-Align instead of RoI-Pooling. 

Following on from Mask R-CNN, PANet \cite{liu2018path} improved the information flow from the backbone to the heads using bottom-up paths in the feature pyramid and adaptive feature pooling. In Mask Scoring R-CNN \cite{huang2019mask}, the network estimates the IoU of the predicted mask and uses it to improve the prediction scores.  HTC \cite{chen2019hybrid} introduced the cascade of masks by merging detection and segmentation features and achieved enhanced detections.  ShapeMask \cite{kuo2019shapemask} introduced class-dependent shape priors and used them as preliminary estimates to obtain the final detection. Instead of building Faster R-CNN, CenterMask \cite{lee2019centermask} built its work on FCOS \cite{tian2019fcos} and applied spatial attention for mask generation. The above methods use the binary-grid representation of masks.

In contrast, PolyTransform \cite{liang2019polytransform} uses a polygon representation and requires a mask for the first stage. The initial mask is refined by a deforming network to obtain the final prediction. These methods accomplish state-of-the-art accuracy, but they are generally slower than one stage methods.

\subsection{One stage instance segmentation}
\quad YOLACT \cite{bolya2019yolact} generated prototype masks and simultaneously produced bounding boxes and combination coefficients. They cropped the prototype masks with the bounding boxes and made a weighted sum of the cropped prototype masks using the combination coefficients to construct the final mask. Likewise, Embedmask \cite{ying2019embedmask} generated pixel embeddings that differentiate each instance in the image and simultaneously produced bounding boxes and proposal embeddings.Here, they form the mask by comparing the proposal embedding with all pixel embeddings in the produced bounding box area. In addition to the previous binary-grid representation approaches, there are a few methods that employ polygon representation. 

ExtremeNet \cite{zhou2019bottom} used keypoint detection to obtain the extreme points of an object. Then a rough mask was created by forming an octagon from the extreme points. Polarmask \cite{xie2019polarmask} performed a dense regression of the distances from the mask center to points on the outer contour in polar coordinates. Additionally, since FCOS \cite{tian2019fcos} showed that the detections near object boundaries were generally inaccurate, they likewise adopted the concept of centerness, which gave greater importance to the detections near the center and enhanced the prediction quality. ESE-Seg \cite{xu2019explicit} trained a network to predict a shape vector that is transformed into contour points in the polar representation. Although their mask representation requires fewer parameters than the other representations, their training method is not optimal, as mentioned in section \ref{sec:intro}. Therefore, we propose employing \textit{differentiable shape decoders} for training, which we explain in the next section.

\section{OUR METHOD}
\label{sec:pagestyle} 
FourierNet is an anchor-free, fully convolutional, single-shot network, and figure \ref{fig:net_arch} illustrates its design. Following its backbone, it has a top-down feature pyramid network (FPN) \cite{lin2017feature} with lateral connections, in which we connect five heads with different spatial resolutions. These heads predict a set of classification scores, centerness, and Fourier coefficients at each spatial location in the feature map. The classification branch predicts scores for each class. Centerness is a term that measures the closeness of a feature point to the mask's center, and we explain it in section \ref{ssec:centerness}. Moreover, we describe the Fourier coefficients in the following mask representation section.

\subsection{Mask representation}
\label{ssec:representation}
\quad FourierNet uses polygon representation to represent masks. The network generates these polygons by a sequence of contour points, represented by polar or Cartesian coordinates. The following two sections describe polar and Cartesian representations, respectively.
\subsubsection{\textbf{Polar}}
\begin{figure}[h]
	\centering
	\includegraphics[width=0.95\linewidth]{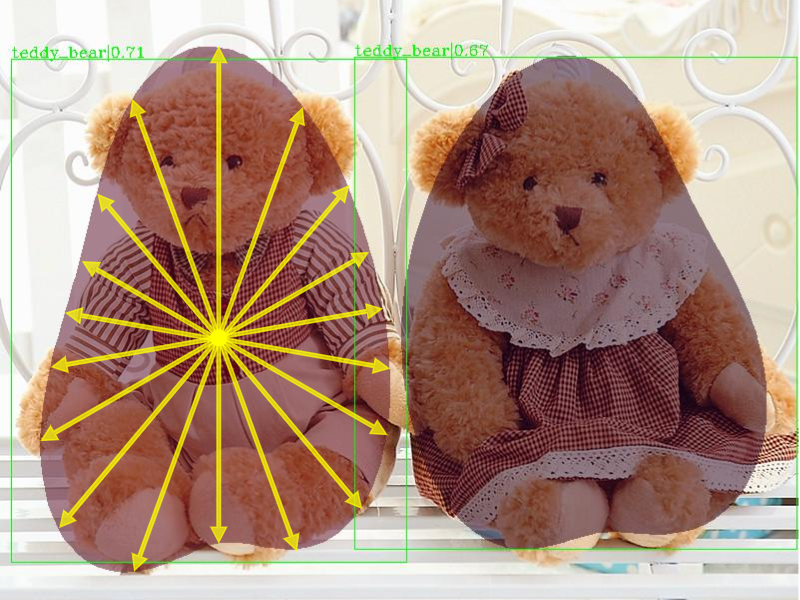}
	\caption{On the left object, \textbf{18 rays} are extended by the lengths $P_i$ from a feature point $i$ (a potential center point). The contour points are the endpoints of these rays. Note that this figure is simplified for illustration purposes only (The actual mask generated in this image has 90 contour points) }
	\label{fig:polar_rep}
\end{figure}

In polar representation, for each feature point $ i $ near the contour's center, we extend $ N $ rays to the point of intersection with the object boundary, as shown in figure \ref{fig:polar_rep}.
 \begin{figure*}[t]
	\centering
	\includegraphics[width=0.99\linewidth]{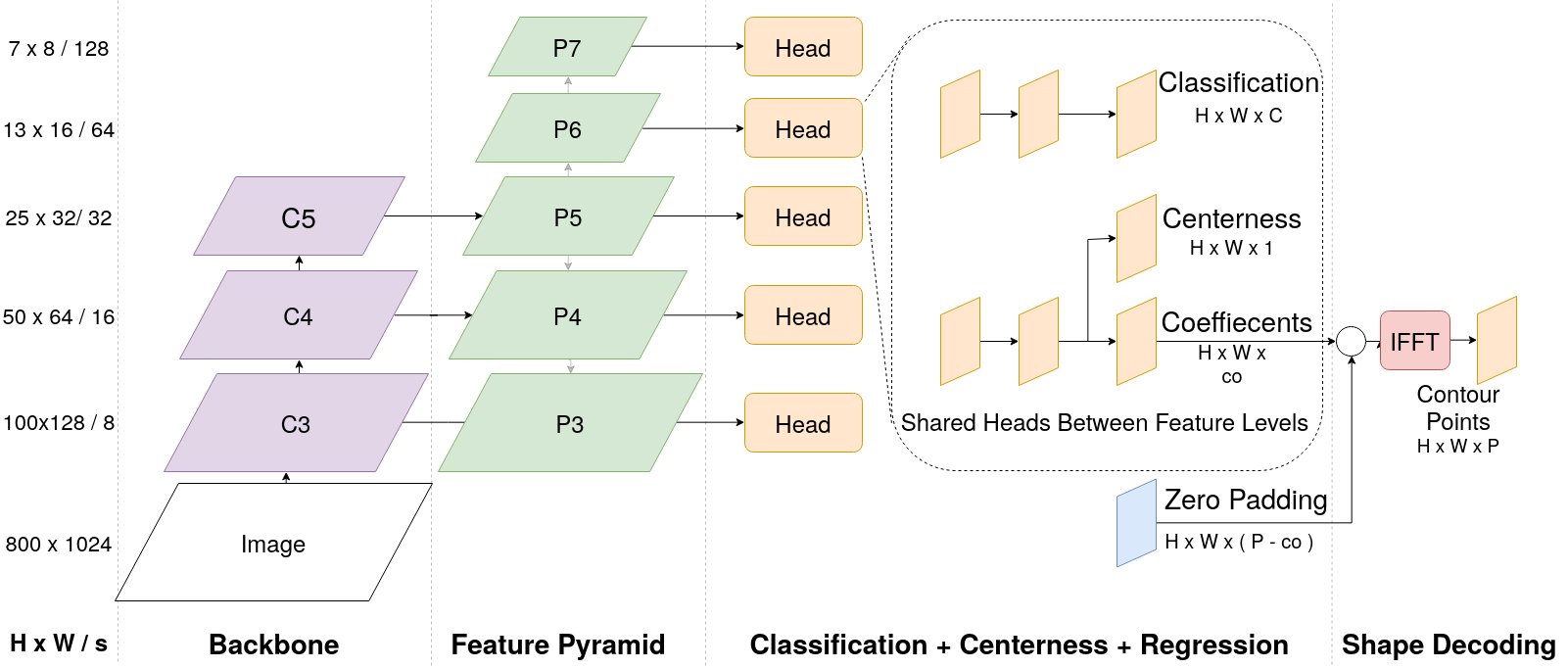}	
	\captionof{figure}{The FourierNet architecture (for polar representation only). FourierNet predicts 5 heads (with shared weights) at various spatial resolutions. More uniquely, each head predicts coefficients of a Fourier series which is converted into contour points using an \textit{Inverse Fast Fourier Transform} (IFFT). Note that IFFT is differentiable and thus termed \textit{differentiable shape decoders}. }
	\label{fig:net_arch}
\end{figure*}
The angle between the rays $\Delta\theta$ is constant and defined by $360^{\circ}/N$. The length of these rays from the center point are described by $P_i=\{p_{0,i},p_{1,i},...,p_{N-1,i}\}$. If there is more than one intersection point, the point with the longest distance is selected. Furthermore, a constant $\epsilon=10^{-6}$ is assigned to rays that do not have intersection points, which occurs when the feature point $i$ is outside or on the contour's boundary. Note that the emerged contour would only approximate the ground truth contour even with a high number of rays; however, the IOU values can reach up to 0.95 \cite{xu2019explicit}.
To determine $P_i$, we apply an \textit{Inverse Fast Fourier Transform} (IFFT) to the coefficients predicted by the network (figure \ref{fig:net_arch}). 
The inverse discrete Fourier transform is defined by 
\begin{equation}
\label{eq:ifft}
p_{n,i} = \frac{1}{N}\sum_{k=0}^{N-1}x_{k,i} e^{\frac{j2\pi kn}{N}},
\end{equation}
where $p_{n,i}$ is the $n_{th}$ ray in $P_i$ and $x_{k,i}$ is the $k_{th}$ coefficient of $X_i$, which is the Fourier transform of $P_i$. In cases where we predict more rays than Fourier coefficients, the network predicts a subset of the coefficients $S_i \subset X_i$ and then we pad the rest of the output tensor with zeros on higher frequency components. This is done to equalize the dimensions before and after the IFFT. Note that the IFFT algorithm is differentiable and therefore the training is done directly on $P_i$ and thus the name \textit{differentiable shape decoder}.
\subsubsection{\textbf{Cartesian}}
\begin{figure}[t]
	\centering
	\includegraphics[width=0.85\linewidth]{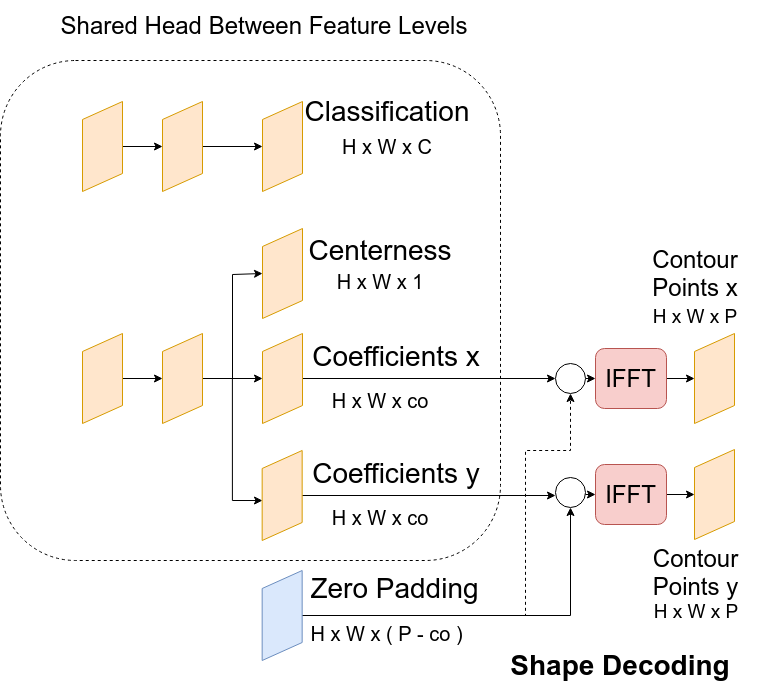}
	\caption{The FourierNet head for Cartesian representation.}
	\label{fig:cart_head}
\end{figure}

Polar representation generates convex masks since, for each angle, there is only one possible ray length. To represent arbitrary masks, we can use Cartesian coordinates. For Cartesian representation, we modified FourierNet head to predict the $x$ and $y$ (Cartesian) coordinates of each contour point of the mask rather than the ray lengths (polar). Figure \ref{fig:cart_head} shows the modified structure of the FourierNet head for Cartesian representation.

Since $x$ and $y$ are independent entities, two separate branches of Fourier coefficients are utilized instead of a single one. An IFFT is applied to each of these branches separately. For Cartesian's case, the $p_{n,i}$ in equation \ref{eq:ifft} refers to the distance of the $n_{th}$ contour point from the $i_{th}$ feature point, in either $x$ or $y$ directions.  For cases where contour points are more than Fourier coefficients, we pad the output tensor with zeros as in polar representation.

\subsection{Centerness}
\label{ssec:centerness}
Centerness is a term that measures the closeness of a feature point to the center of a mask, and it used to filter out weak detections \cite{tian2019fcos}. We utilize \textit{polar centerness} in the case of polar representation and \textit{Gaussian centerness} in the case of Cartesian representation.  Both are detailed in the following sections, respectively.
\subsubsection{\textbf{Polar centerness}}
Polar Centerness (PC) \cite{xie2019polarmask} is defined for the $i_{th}$ feature point as 
\begin{equation}
\label{eq:polar_cent}
PC_i = \sqrt{\frac{\min({p_{0,i},p_{1,i},...,p_{N-1,i}})}{\max({p_{0,i},p_{1,i},...,p_{N-1,i}})}},
\end{equation}

where $p_{n,i}$ are the ray lengths. During inference, we multiply this value with the classification score to keep the locations, which could produce the best detection. 

We argue that this metric would be low if the object's mask shape is not circular, and since we multiply it by the classification score, it will lower the probability of predicting such objects. PolarMask introduced a hyperparameter called \textit{Centerness Factor} (CF) to overcome this problem, which is added to the centerness to increase its value. To the best of our knowledge, this offset defeats the purpose of centerness, since it artificially raises confidence and sometimes even exceeds 1. Moreover, it does not explicitly solve the problem of low centerness of non-circular objects. Therefore, we introduce \textit{Normalized Centerness} ($NC$) which is defined for a feature point $i$ by
\begin{equation}
NC_i = \frac{PC_i}{PC_{\max}},
\end{equation}
where $PC_{max}$ is the polar centerness of the \textit{center of mass} of an instance. The maximum value of the $NC_i$ is clamped to one, when the center of mass does not have the highest polar centerness value.
\subsubsection{\textbf{Gaussian centerness}}
In the case of centerness for Cartesian representation, we can not adopt equation \ref{eq:polar_cent} directly. Accordingly, we apply a Gaussian distribution to represent the probability of a point being at the object's center. For the $ith$ feature point having the location ($m$,$n$) in the feature map, Gaussian centerness (GC) is defined as
\begin{equation}
\label{eq:gauss_cent}
GC = e^{-\alpha(\frac{m-\mu_x}{\sigma_x})^2} e^{-\alpha(\frac{n-\mu_y}{\sigma_y})^2},
\end{equation}

where $\mu_x$ and $\mu_y$ are the means (center points), and $\sigma_x$ and $\sigma_y$ are the standard deviations of a mask instance in $x$ and $y$ directions respectively and $\alpha$ is a hyperparameter used for controlling the decaying rate. Note that we multiply the two Gaussians, which enforces a probability of 1 only if both $m$ and $n$ are at the object's center. On all the other locations, the decaying functions' product reduces the centerness depending upon the standard deviation in both $x$ and $y$ directions of the mask instance. Notice that GC solves the problem of low centerness for non-circular objects, and therefore centerness factor can be completely avoided. 
\subsection{Loss function}
The overall loss function comprises of four components, which is defined as:
\begin{equation}
\label{eq:loss}
L_{total} = L_{cls} + L_{box} + L_{cent} + L_{mask}.
\end{equation}
We use \textit{focal loss} \cite{lin2017focal} for the classification loss $L_{cls}$ and  \textit{IOU loss} \cite{yu2016unitbox} for the bounding box loss $L_{box}$. Note that bounding box branch is an optional branch and therefore not explicitly shown in the figure \ref{fig:net_arch}. For centerness loss $L_{cent}$, we employ \textit{binary cross entropy} for both \textit{polar centerness} and \textit{Gaussian centerness}. For mask loss $L_{mask}$, we utilize two different loss functions for polar and Cartesian representations. For polar representation, we adopt \textit{polar IOU loss} from \cite{xie2019polarmask}. For Cartesian representation, we employed both \textit{smooth L1} loss and \textit{Chamfer distance} loss \cite{fan2017point}. In the following section, Chamfer distance loss has been explained.
\subsubsection{\textbf{Chamfer distance loss}}
\label{sssec:chamfer}
To train the ($x$, $y$) contour points, Chamfer distance loss \cite{fan2017point} is adopted. It is defined as
\begin{equation}
\label{eq:chamfer}
CD = \sum_{\substack{a\in S_1}}\min_{\substack{b\in S_2}} \parallel D(a,b) \parallel_2^2 + \sum_{\substack{b\in S_2}}\min_{\substack{a\in S_1}} \parallel D(a,b) \parallel_2^2,
\end{equation}
where $S_1$ and $S_2$ are the sets of predicted contour points and ground truth contour points respectively, $a$ and $b$ are elements (individual contour points $(x,y)$) of the sets $S_1$ and $S_2$ respectively and $D(a,b)$ is the euclidean distance between any two points $a$ and $b$ respectively. Please note that the centroid of the object is taken as reference for the contour points. We normalize the Chamfer distance by dividing it by the average of the height and width of the ground truth bounding box. Without the normalization, Chamfer distance becomes exceptionally large which leads to overflows and exploding gradients. Moreover, normalization avoids the problem of manually weighing classification, centerness and mask losses. 

Chamfer loss employ the nearest neighbor approach to assign predictions to targets. There is a potential for a mismatch between predictions and targets if they are assigned index-wise, since there could be a positional offset between their starting indices even when the masks look visually the same. Due to nearest neighbor assignment, Chamfer loss overcomes this risk. Moreover, in Chamfer distance loss, nearest neighbors of both predictions and targets are considered separately. If we take the nearest neighbors of all the predicted points only, some target points are overlooked while training, and vice versa. This leads to some overlooked sharp edges in targets and some under-trained weights predicting inaccurate contours. The dual terms in Chamfer loss avoids this problem.

\section{Experiments}
\label{sec:exp}

\quad We conducted the experiments on the COCO 2017 benchmark \cite{lin2014microsoft} divided into 118K training and 5K validation data splits. We based our work on PolarMask \cite{xie2019polarmask} implementation, which uses the mmdetection framework \cite{mmdetection}. Unless otherwise stated, we did all the experiments using a  pre-trained ResNet-50 \cite{He2015} on ImageNet \cite{krizhevsky2012imagenet}. We trained the networks for 12 epochs with an initial learning rate of 0.01 and a mini-batch of 4 images. The learning rate was reduced by a factor of 10 at epochs 8 and 11. We used stochastic gradient descent (SGD) with momentum (0.9) and weight decay (0.0001) for optimization. We resized the input images to 1280$\times$768 pixels.
\begin{figure*}[h]
	\centering
	\begin{subfigure}{0.49\linewidth}
		\centering
		\includegraphics[width=0.95\linewidth]{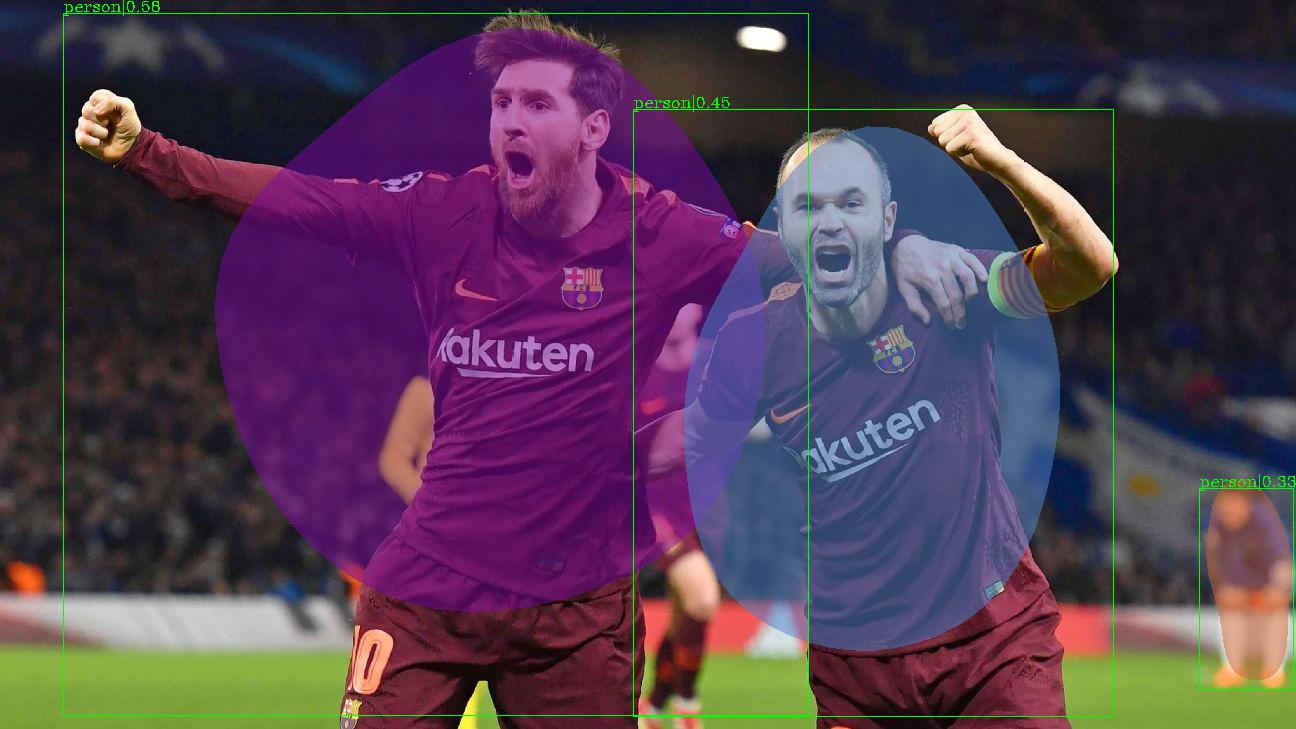}
		\caption{8 coefficients (Smooth L1)}
		\label{fig:8_L1_messi}
	\end{subfigure}
	\begin{subfigure}{0.49\linewidth}
		\centering
		\includegraphics[width=0.95\linewidth]{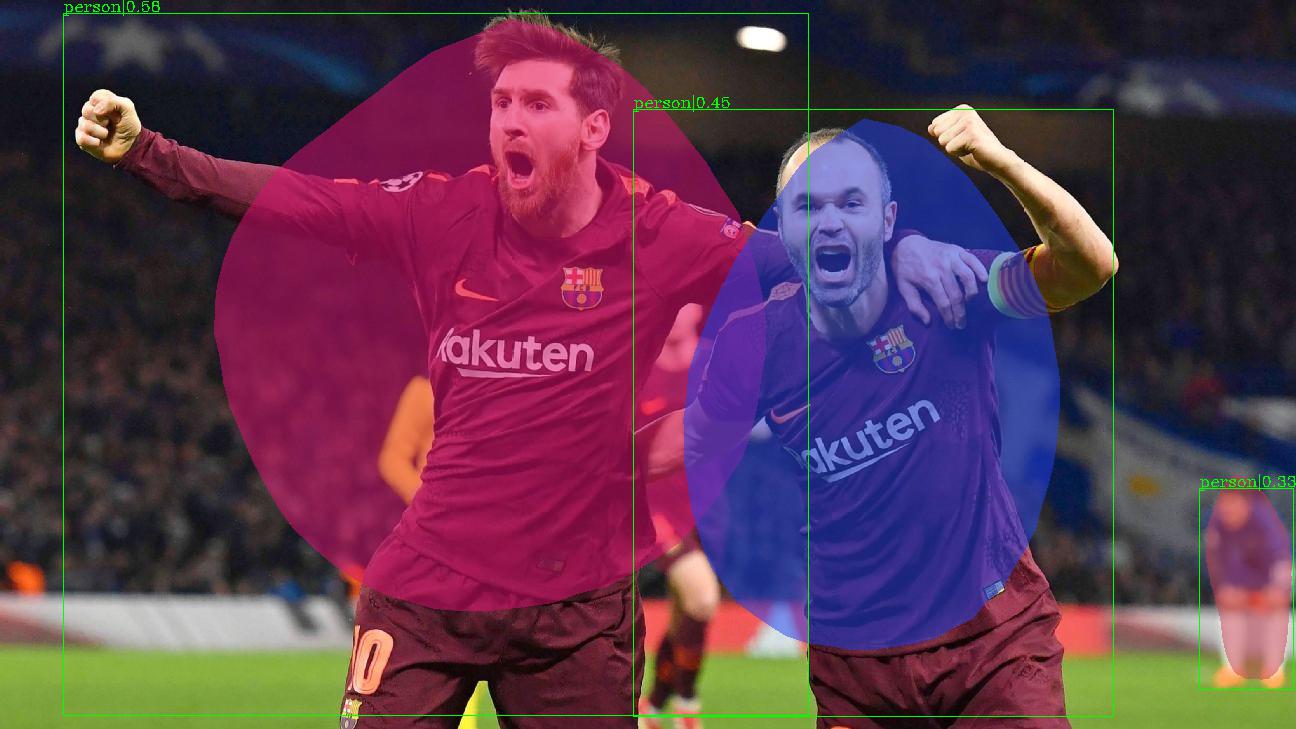}
		\caption{36 coefficients (Smooth L1)}
		\label{fig:36_L1_messi}
	\end{subfigure}
	\begin{subfigure}{0.49\linewidth}
		\centering
		\includegraphics[width=0.95\linewidth]{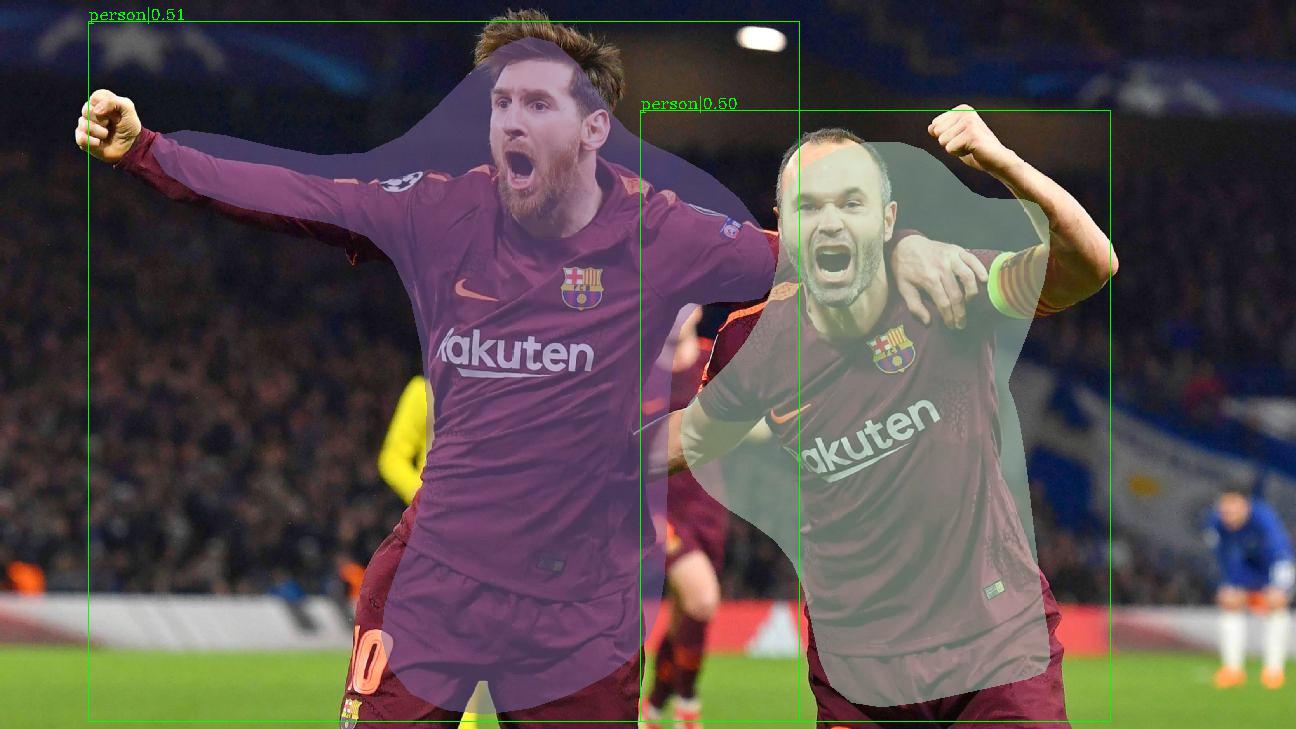}
		\caption{8 coefficients (Chamfer distance)}
		\label{fig:cart_8}
	\end{subfigure}
	\begin{subfigure}{0.49\linewidth}
		\centering
		\includegraphics[width=0.95\linewidth]{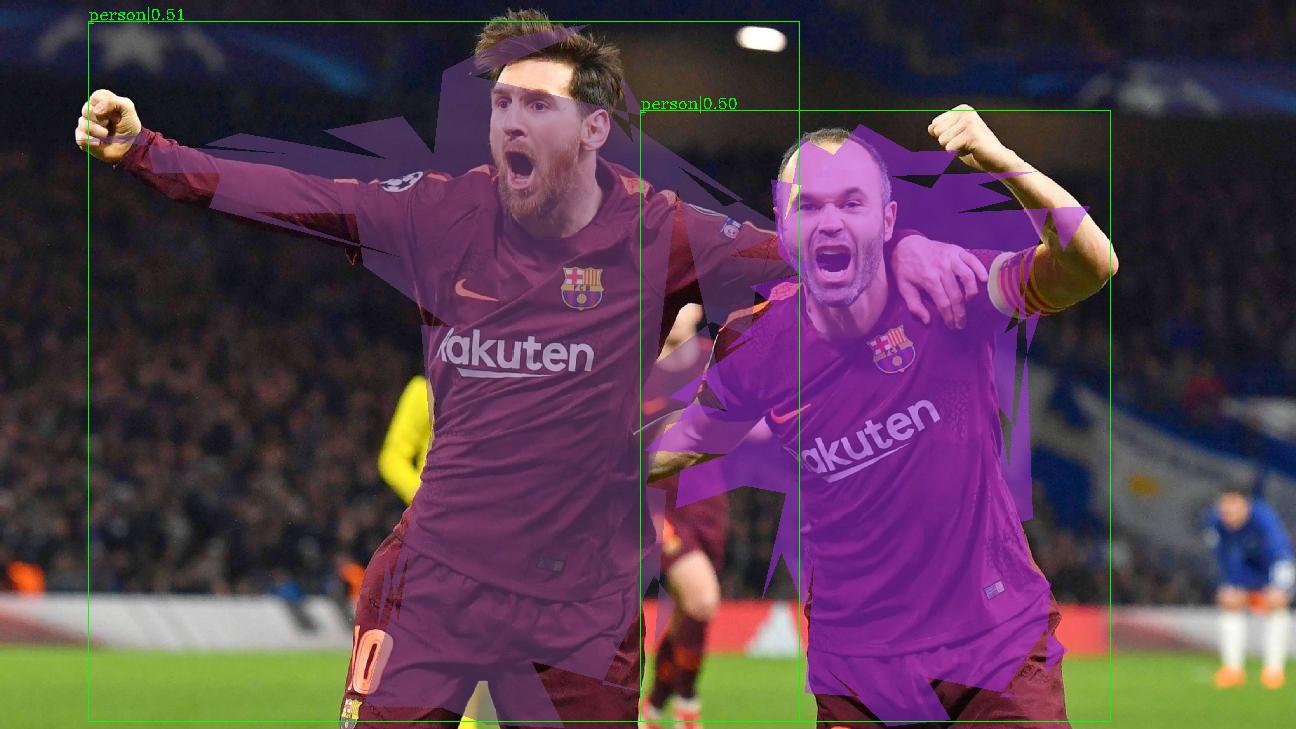}
		\caption{36 coefficients (Chamfer distance)}
		\label{fig:cart_36}
	\end{subfigure}

	\caption{Examples of the network using Cartesian coordinates. Sub-figure \ref{fig:8_L1_messi} and \ref{fig:36_L1_messi} are trained using smooth L1 loss. Sub-figure \ref{fig:cart_8} and \ref{fig:cart_36} are trained using Chamfer distance loss.}
	\label{fig:cart_compare}	
\end{figure*}
 We employed \textit{mean average precision} (mAP) as our evaluation criteria. We took the average of the area under the \textit{precision-recall} curve on 10 \textit{intersection over union} (IoU) levels between 0.5 and 0.95 (step size 0.05) and then averaged over all 80 categories in COCO 2017 dataset.
\subsection{Cartesian representation vs. polar representation}
\label{ssec:cartvspolar}
Table \ref{tab:cartesian_compare} shows a comparison between various networks trained on Cartesian representation using \textit{smooth L1 loss} and \textit{Chamfer distance loss}. Two networks trained on polar representation using \textit{polar IOU loss} are also shown for comparison. The training set up and hyper-parameters are the same as described above. The network with smooth L1 loss performs worst with 13.6 mAP using 8 coefficients. Since smooth L1 loss associates predictions and targets index-wise, there is a mismatch between associations (as discussed in section \ref{sssec:chamfer}). The model ignores the edges and learns the average mask (ellipse-like), as shown in figures \ref{fig:8_L1_messi} and \ref{fig:36_L1_messi}. Moreover, the difference between mask quality of 8 and 36 coefficients is insignificant, as visible in the figures.

\begin{table}[h]
	\centering
	\begin{tabular}{c|cc|c}
		Coeff. & \shortstack{Loss\\mask} & \shortstack{Loss\\centerness}& mAP   \\
		\hline

		8  & Smooth L1 & Gaussian & 13.6    \\
		36  & Smooth L1 & Gaussian & 13.5    \\
		\hdashline
		8  & Chamfer & Gaussian & 22.9   \\
		36  & Chamfer & Gaussian & 22.4   \\
		\hdashline
		8  & Polar IOU & Polar & 26.8  \\
		36  & Polar IOU & Polar & 28.0  \\		
		\hline
	\end{tabular}
	\caption{Comparison of mAP for the Cartesian and polar representations}
	\label{tab:cartesian_compare}
\end{table}
\begin{figure}[h]
	\centering
	\includegraphics[width=0.95\linewidth]{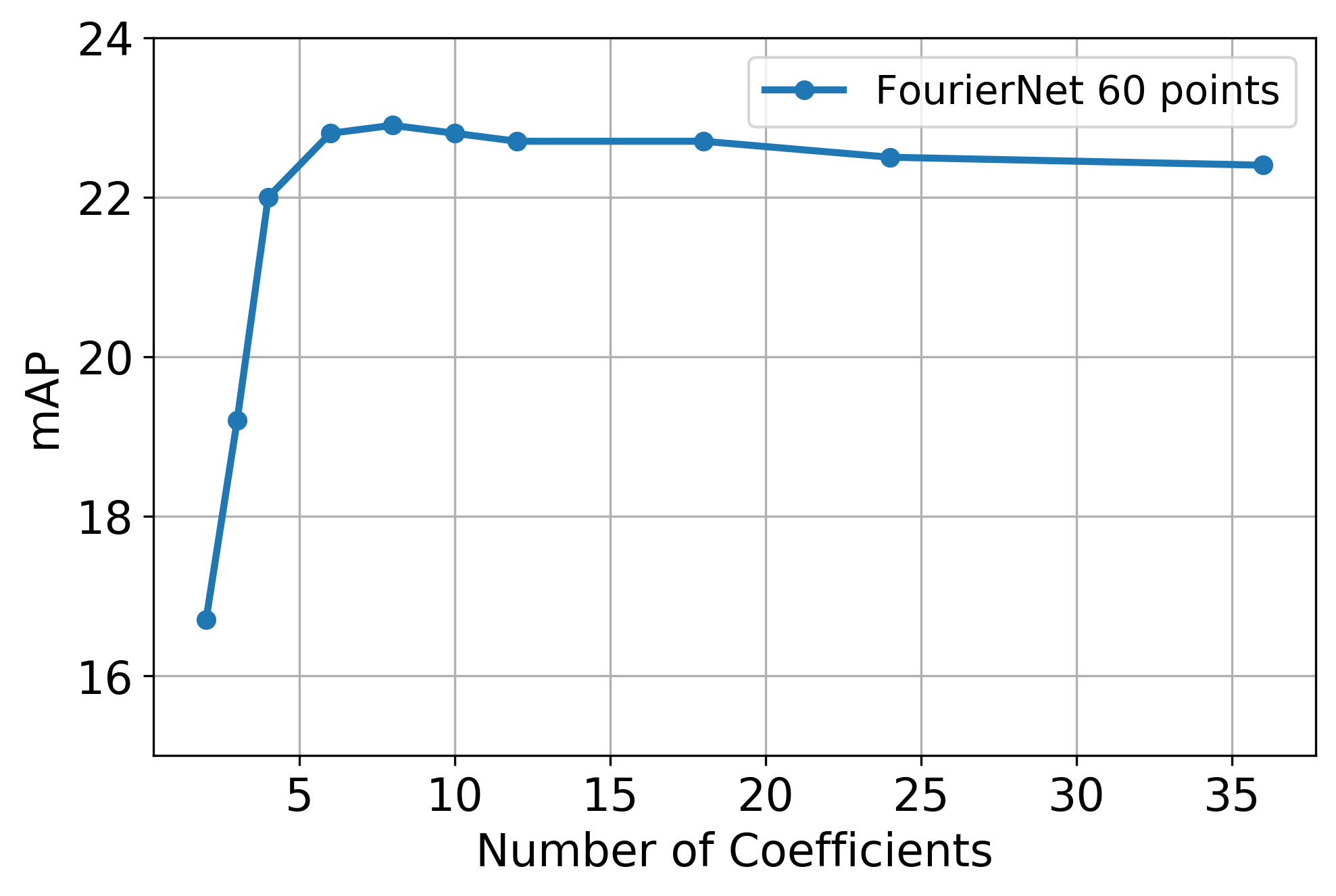}
	\caption{Evolution of performance of FourierNet in Cartesian representation with changing number of coefficients trained with Chamfer distance loss.}
	\label{fig:map_vs_cart}
\end{figure}
\begin{figure*}[h]
	\begin{subfigure}{.33\linewidth}
		\centering
		\includegraphics[width=0.95\linewidth]{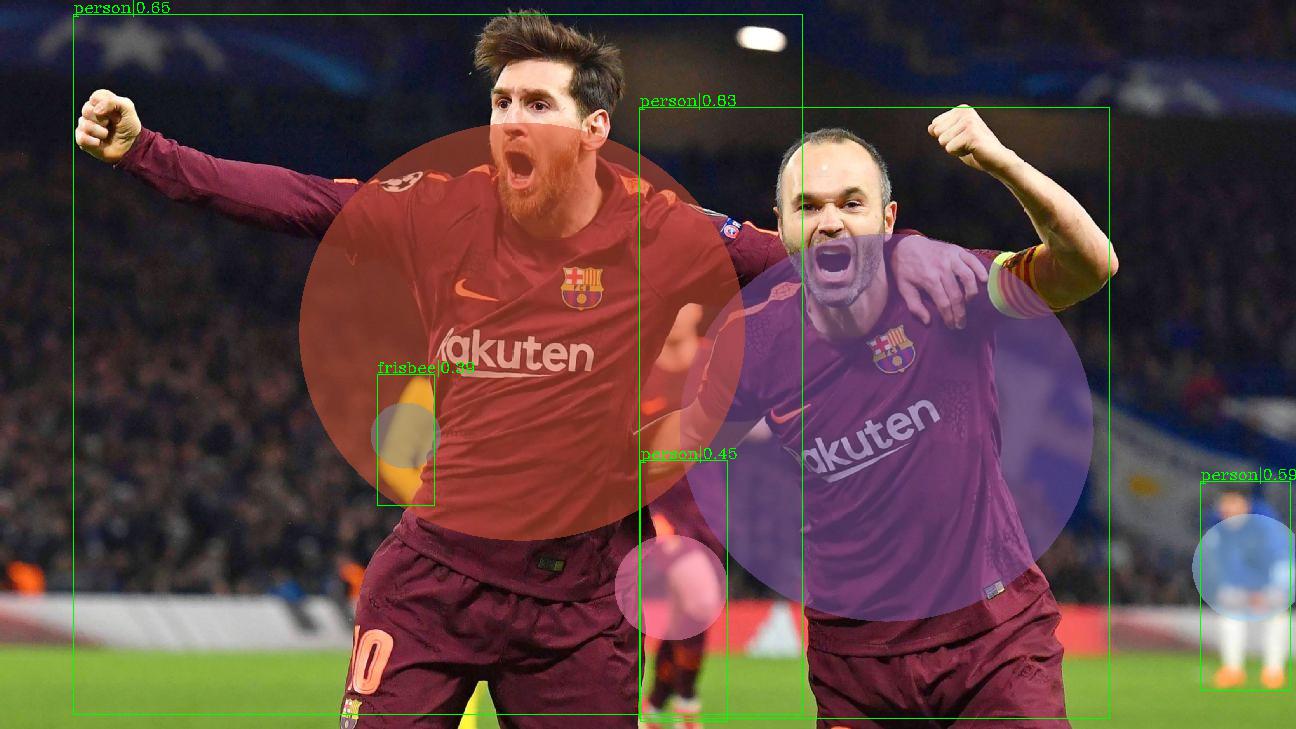}
		\caption{2 coeff. (4 parameters.)}
		\label{fig:messi_2}
	\end{subfigure}
	\begin{subfigure}{.33\linewidth}
		\centering
		\includegraphics[width=0.95\linewidth]{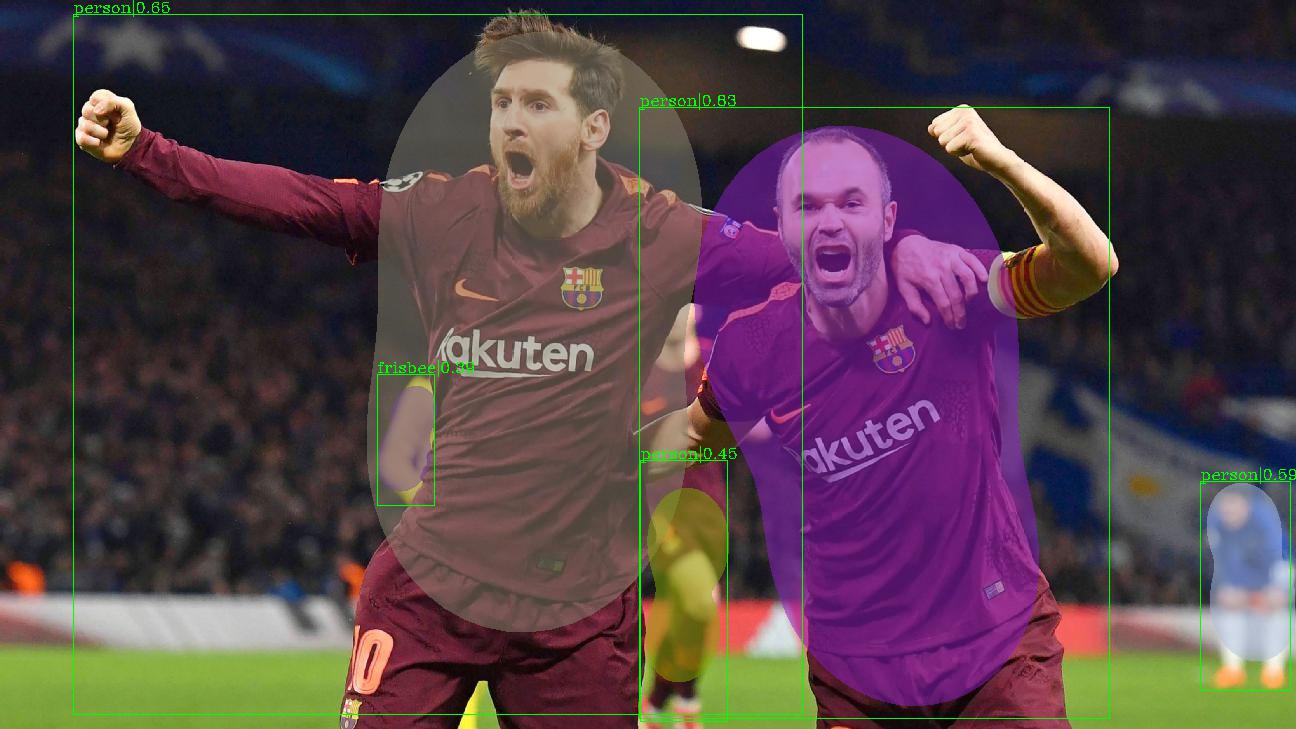}
		\caption{3 coeff. (6 parameters.)}
		\label{fig:messi_3}
	\end{subfigure}
	\begin{subfigure}{.33\linewidth}
		\centering
		\includegraphics[width=0.95\linewidth]{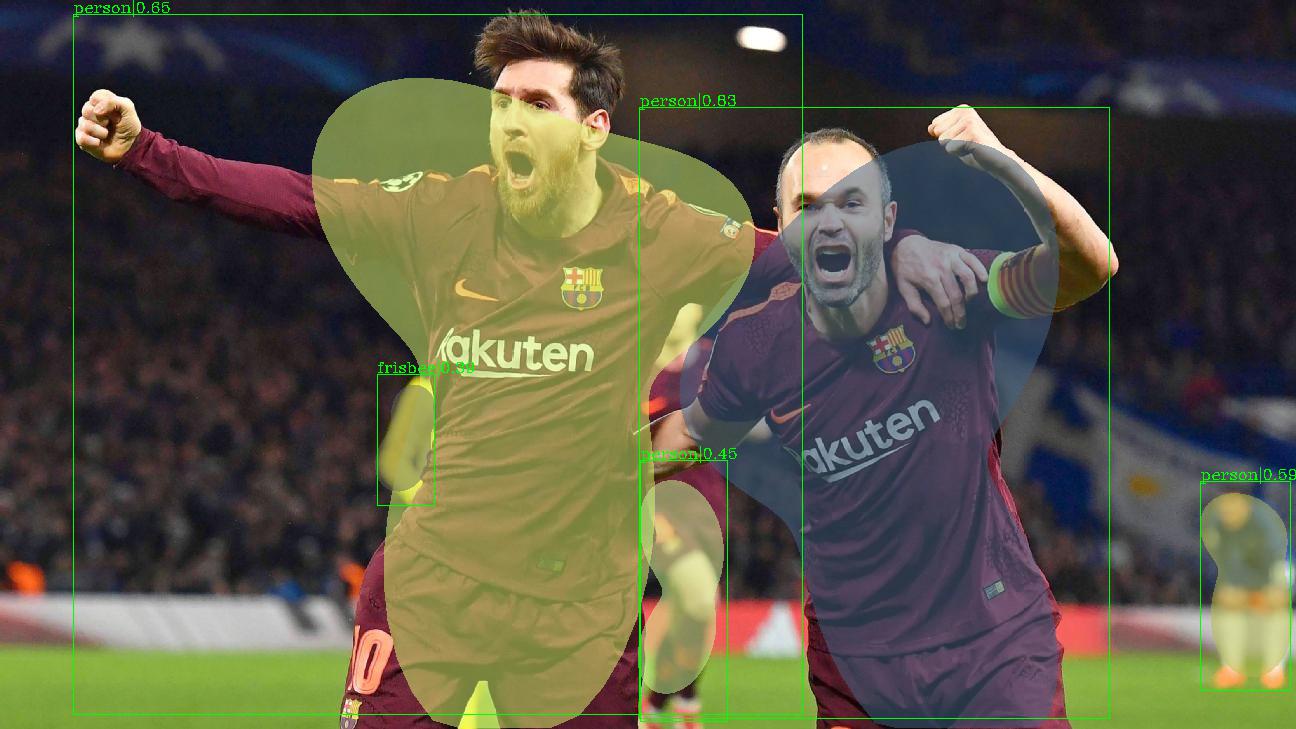}
		\caption{4 coeff. (8 parameters.)}
		\label{fig:messi_4}
	\end{subfigure}
	\begin{subfigure}{.33\linewidth}
		\centering
		\includegraphics[width=0.95\linewidth]{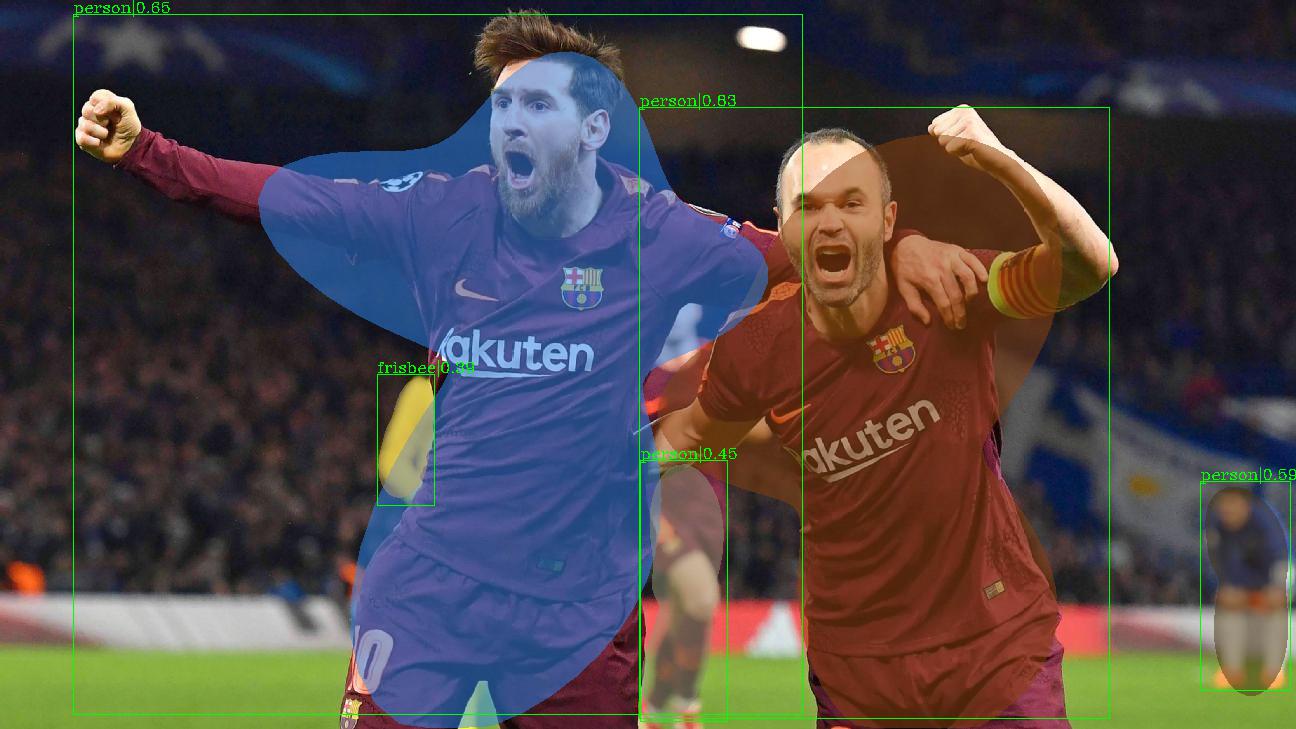}
		\caption{6 coeff. (12 parameters.)}
		\label{fig:messi_6}
	\end{subfigure}
	\begin{subfigure}{.33\linewidth}
		\centering
		\includegraphics[width=0.95\linewidth]{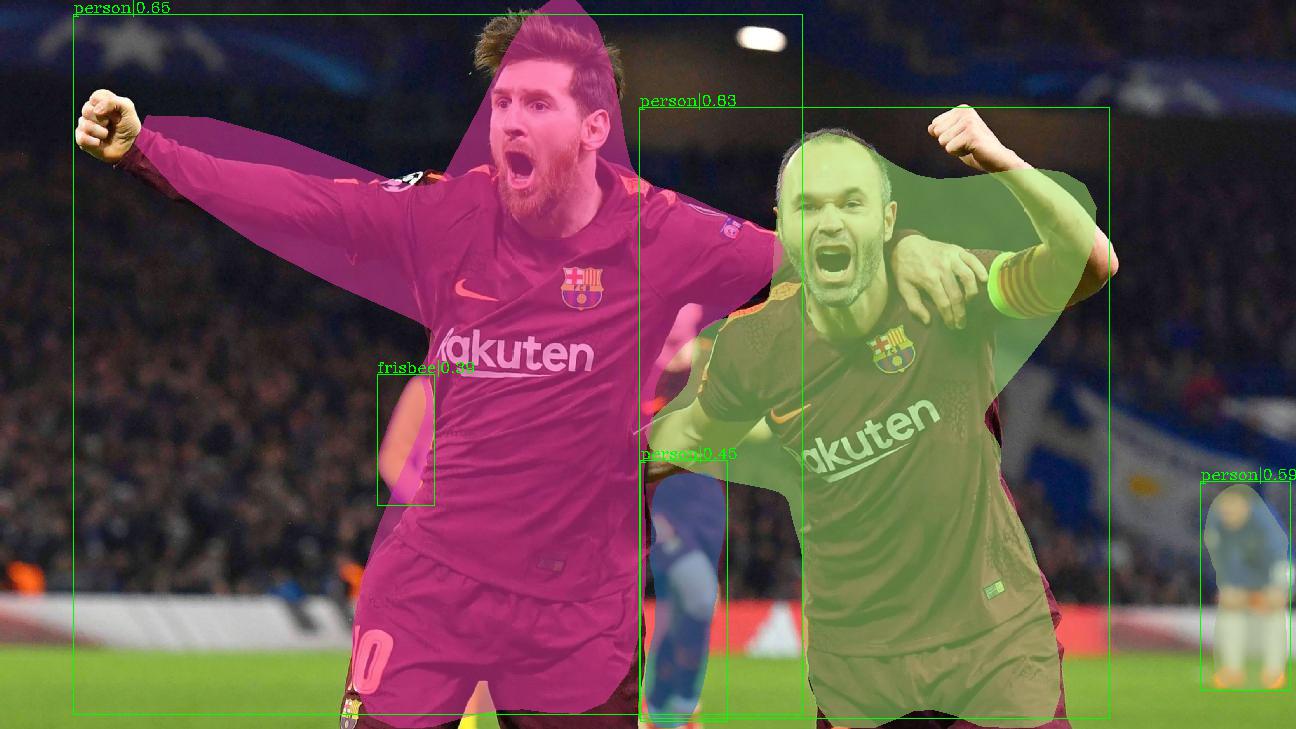}
		\caption{12 coeff. (24 parameters.)}
		\label{fig:messi_12}
	\end{subfigure}
	\begin{subfigure}{.33\linewidth}
		\centering
		\includegraphics[width=0.95\linewidth]{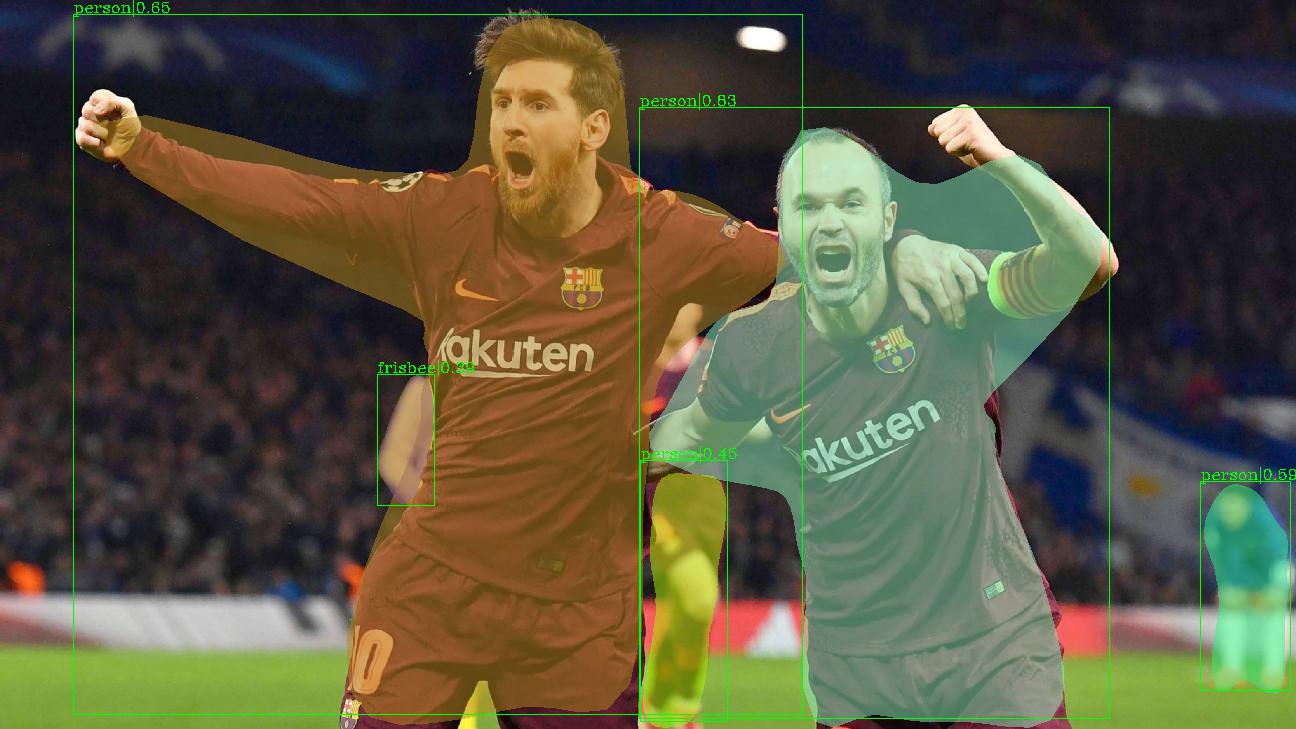}
		\caption{36 coeff. (72 parameters.)}
		\label{fig:messi_36}
	\end{subfigure}
	\caption{Comparison between the predicted mask of using varying number of Fourier coefficients using polar representation.}
	\label{fig:coeff_compare}
\end{figure*}
The networks trained with the Chamfer distance loss, were first pre-trained for one epoch on smooth L1 loss as a warm-up. This provides a good initialization since Chamfer loss greatly benefits from elliptical predictions at start. In both the  experiments, $\alpha=10$ is used in Gaussian centerness (equation \ref{eq:chamfer}) which provides a reasonable balance between a high probability for a point at the center of the object and low values at the mask edges. The network with 8 coefficients using Chamfer distance loss produces the best results with 22.9 mAP in the Cartesian domain. 

Figure \ref{fig:map_vs_cart} shows the evolution of mAP with respect to number of coefficients of Fourier series. It can be seen that the maximum mAP is reached when using 8 coefficients only. Figures \ref{fig:cart_8} and \ref{fig:cart_36} illustrate the masks generated by the network using Cartesian representation. Counter-intuitively, the masks with 8 coefficients are smoother and have a better IOU than the masks with 36 coefficients. As seen in the figure \ref{fig:cart_36}, when using 36 coefficients, there are undesired oscillations which makes the contour worse. One possible reason could be that the gradients are very low for the the higher frequency coefficients during training because they affect the output very little. This eventually leads to under-trained higher frequency coefficients which show erratic results when visualized. Similar networks trained using polar representation show significantly better performance with 26.8 and 28.0 mAP for 8 and 36 coefficients respectively. Therefore, polar representation has been employed in the experiments later in the paper.

\subsection{Ablation study}
\label{sec:ablstd}
All the experiments done in this ablation study section adopt polar representation for masks.
\subsubsection{\textbf{Number of contour points}}
\label{ssec:npoints}
\begin{figure}[h]
	\centering
	\includegraphics[width=0.98\linewidth]{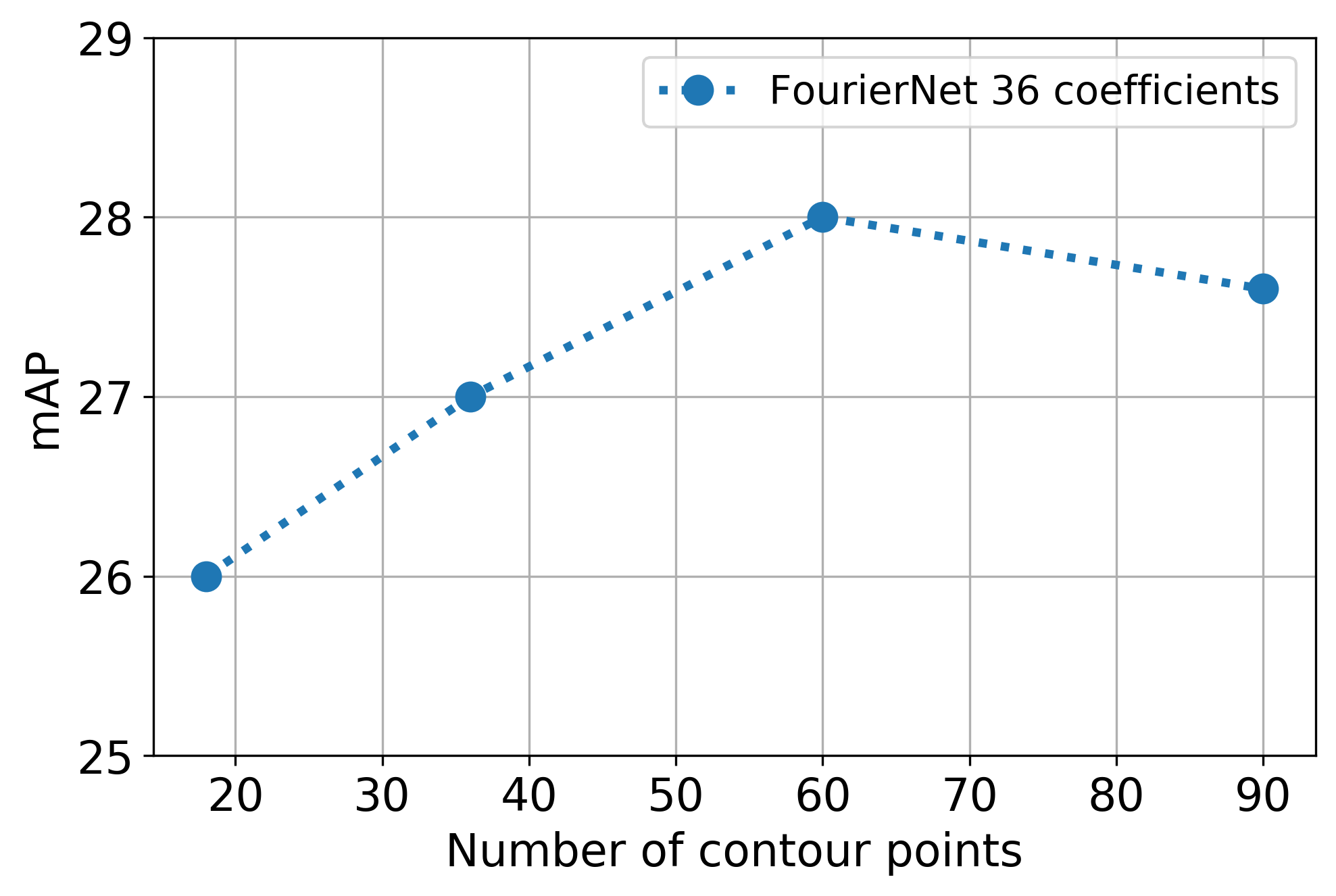}
	\captionof{figure}{The FourierNet in this experiment has 36 coefficients of Fourier series and Resnet-50 backbone. }
	\label{fig:mapvspoints}
\end{figure}

We trained multiple FourierNets were having 18, 36, 60, and 90 contour points and 36 complex coefficients (72 parameters). Figure \ref{fig:mapvspoints} shows that more contour points generally lead to a higher mAP until it reaches a maximum, and then the performance deteriorates.
The FourierNet with 90 points has a lower mAP (27.6) than FourierNet with 60 points (28.0), and a possible reason could be that the added complexity (in terms of contour points) makes the problem harder for the optimizer to learn. Furthermore, while more contour points seem more appealing for large and complicated masks, for smaller objects, it means adding unwanted complexity, which could lower $AP_s$ if not learned correctly, leading to an overall negative effect on performance.
\begin{table*}[t]
	\centering
	\begin{tabular}{c|c|c|c|ccc|ccc|cc}
		Method & B.Bone & Rep. & Param. & mAP & $\textrm{AP}_{50}$ & $\textrm{AP}_{75}$ & $\textrm{AP}_{S}$ & $\textrm{AP}_{M}$ & $\textrm{AP}_{L}$ & FPS & GPU  \\
		\hline
		\textit{two stage} & & & & & & & & & & & \\
		Mask RCNN \cite{he2017mask} & RX-101  & binary grid & 784 & 37.1  & 60.0  & 39.4  & 16.9 & 39.9 & 53.5 & 5.6 & 1080Ti \\
		PANet \cite{liu2018path} & RX-101 & binary grid & 784 & \textbf{42.0} & \textbf{65.1} & \textbf{45.7} & 22.4 & \textbf{44.7} & \textbf{58.1} & - & - \\
		HTC \cite{chen2019hybrid} & RX-101 & binary grid & 784 & 41.2 & 63.9 & 44.7 & \textbf{22.8} & 43.9 & 54.6 & 2.1 & TitanXp \\
		\hline
		\textit{one stage} & & & & & & & & & & &\\
		ESE-Seg-416 \cite{xu2019explicit} & DN-53 & shp. encoding  & 20 & 21.6 & \textbf{48.7} & 22.4 & - & - & - & \textbf{38.5} & 1080Ti\\
		\textbf{FourierNet-640} & R-50 & shp. encoding & 20 & \textbf{24.3} & 42.9 & \textbf{24.4} & 6.2 & 25.9 & 42.0 & 26.6 & 2080Ti \\		
		\hdashline
		ExtremeNet \cite{zhou2019bottom} & HG-104 & polygon & \textbf{8} & 18.9 & 44.5 & 13.7 & \textbf{10.4} & 20.4 & 28.3 & 3.1 & -  \\
		\textbf{FourierNet} & RX-101 & shp. encoding & \textbf{8} & \textbf{23.3}& \textbf{46.7} & \textbf{21.1} & 10.3 & \textbf{25.2} & \textbf{34.4} & \textbf{6.9} & 2080Ti \\
		\hdashline				
		EmbedMask \cite{ying2019embedmask} & R-101 & binary grid & $\dagger$ & \textbf{37.7} & \textbf{59.1} & \textbf{40.3} & \textbf{17.9} & \textbf{40.4} & \textbf{53.0} & 13.7 & V100\\
		YOLACT-700 \cite{bolya2019yolact} & R-101 & binary grid & $\dagger$ & 31.2 & 50.6 & 32.8 & 12.1 & 33.3 & 47.1 & \textbf{23.4} & TitanXp\\
		PolarMask \cite{xie2019polarmask} & RX-101  & polygon & 36 & 32.9  & 55.4  & 33.8  & 15.5 & 35.1 & 46.3 & 7.1* & 2080Ti \\		
		\textbf{FourierNet} & RX-101 & shp. encoding & 36 & 30.6  & 50.8 & 31.8 & 12.7 & 33.7 & 45.2 & 6.9 & 2080Ti \\
		\hline
	\end{tabular}
	\caption{Comparison with state-of-the-art for instance segmentation on COCO test-dev. $\dagger$ The number of parameters are dependent on the size of the bounding box cropping the pixel embedding or mask prototype. * speed tested on our machines.
	}
	\label{tab:sota}
\end{table*}

\subsubsection{\textbf{Number of coefficients}}
\label{ssec:ncoe}
\begin{figure}[t]
	\centering
	\includegraphics[width=0.98\linewidth]{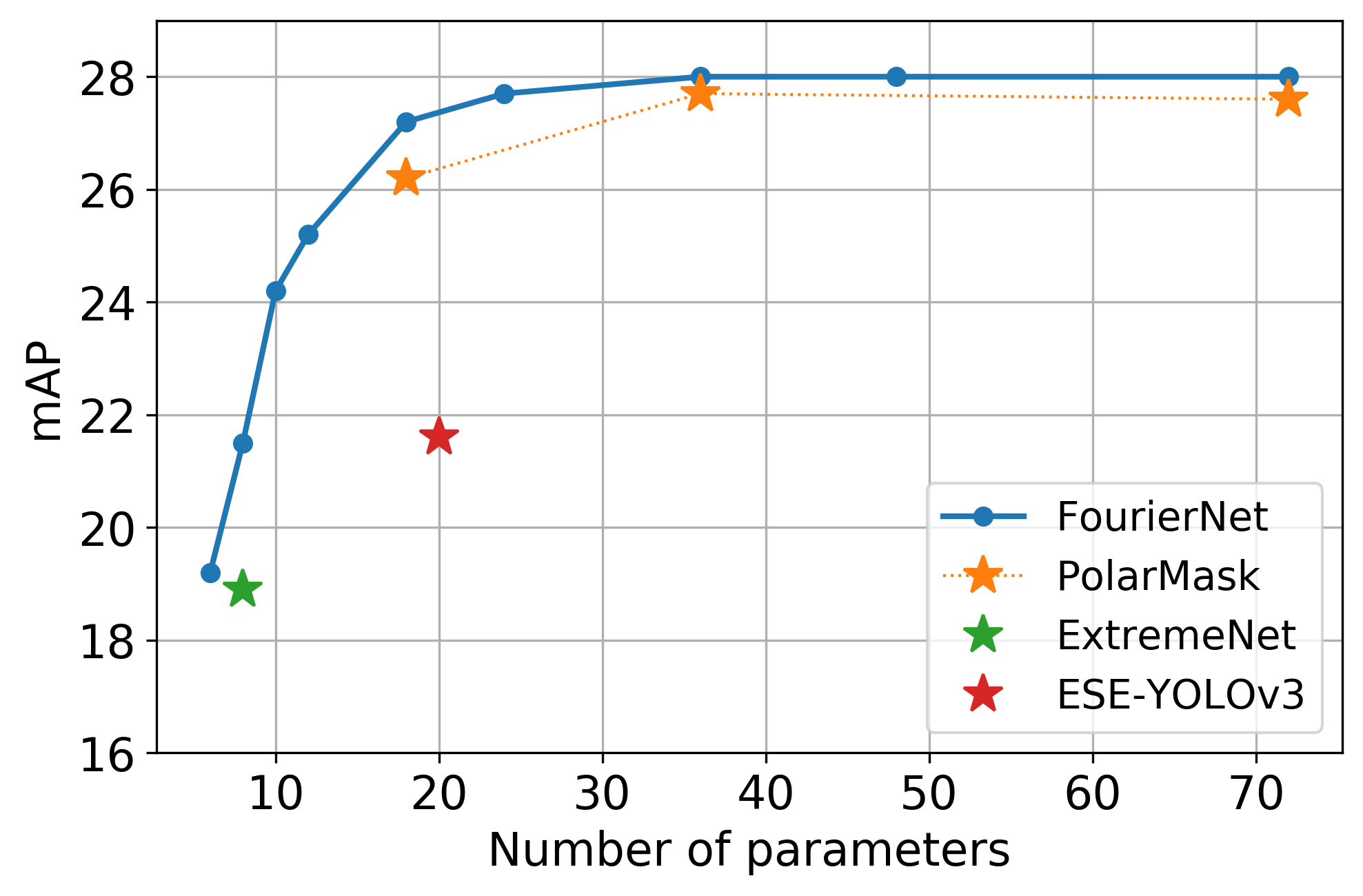}
	\captionof{figure}{The FourierNet in this experiment has a \textbf{Resnet-50}  backbone and \textbf{60 contour points}. The number of parameters are the complex (real+imaginary) coefficients of the Fourier series i.e 36 coefficients = 72 parameters. The Polarmask used in this experiment has the same backbone as well.}
	\label{fig:coevsap}
\end{figure}

From the results of the previous section \ref{ssec:npoints}, we choose a FourierNet with 60 contour points for this study. Figure \ref{fig:coevsap} illustrates the progression of the accuracy of the FourierNet for a varying number of parameters. We generated the curve by testing the network multiple times. Each point in the curve refers to a test wherever we use a subset of the network output tensor with the lowest frequency coefficients. The rest of the higher frequency coefficients were replaced with zeros to inhibit their effects (explained in section \ref{ssec:representation}).

As the number of parameters increases, the mAP sharply increases until around 18 parameters, and after 36 parameters, it saturates. We also observed that the FourierNets with 18, 36, and 90 points showed the same trend of this curve as FourierNet with 60 points, so we did not plot them. Furthermore, we visualize in figure \ref{fig:coeff_compare}, the network's output with varying number of suppressed higher frequencies. We obtain smooth contours if we use a low number of coefficients, and when we utilize only two coefficients (figure \ref{fig:messi_2}), all the predictions become ellipses. For 36 coefficients (figure \ref{fig:messi_36}), we acquire a reasonably good prediction, with some limitations in the non-convex regions of the mask due to polar representation.  

\subsubsection{\textbf{Coefficients regression (CR) vs. differentiable shape decoding (DSD)}}
\label{ssec:crvsdsd}

\quad The lower frequency coefficients of a Fourier series have a higher impact on the contour, which we can infer from the experiments in section \ref{ssec:npoints}. However, unweighted coefficient regression focuses equally on all coefficients during training, which is not optimal for shape decoders. On the contrary, when trained on contour points, the optimizer can inherently learn to prioritize the Fourier series' lower frequency coefficients and achieve \textit{automatic weight balancing}.
\begin{table}[h]
	\centering
	\begin{tabular}{c|ccc}
		Method & mAP & $\textrm{AP}_{50}$ & $\textrm{AP}_{75}$ \\
		\hline
		Coefficient Regression & 5.3  & 14.9  & 3.1  \\
		Differentiable Shape Decoding & 26 & 46.6  & 25.8  \\
		\hline
	\end{tabular}
	\caption{Coefficients regression (CR) vs. Differentiable shape decoding (DSD)}
	\label{tab:crvsdsd}
\end{table} 

To verify this hypothesis, we trained a network with 18 coefficients and directly regressed the coefficients using a smooth L1 loss. It attained a mAP of 5.3 (table \ref{tab:crvsdsd}), which is weak compared to a similar network trained on contour points (26 mAP from figure \ref{fig:coevsap}) and it validates our initial intuition. Moreover, the qualitative results of CR showed out of size masks, which is a sign of errors in low-frequency coefficients.

ESE-Seg \cite{xu2019explicit} also used CR and compared various function approximators. They reported the best performance on Chebyshev polynomials and argued that they have the best distribution. However, we argue that if they had used optimized weights for Fourier coefficients during training, they would reach better performance. We verified with our results that DSD is better than CR because it does automatic weight balancing.

\subsubsection{\textbf{Polar centerness (PC) vs. normalized centerness (NC)}}
\label{ssec:pcvsnpc}

\quad We trained two networks with 60 contour points and 36 coefficients on NC and PC. From the results in table \ref{tab:pcvsnpc}, we perceive that NC is better than PC when we set the centerness factor (CF) to zero, which means that it is generally a more reliable centerness metric. However, to obtain the best performance, we still need to use the CF hyperparameter.
\begin{table}[h]
	\centering
	\begin{tabular}{c|c|c|cc}
		Centerness & CF & mAP & $\textrm{AP}_{50}$ & $\textrm{AP}_{75}$ \\
		\hline
		Polar & 0 & 26.3  & 42.8  & \textbf{27.7}   \\		
		Normalized & 0 & \textbf{27.0} & \textbf{47.8} &  26.9   \\
		\hdashline
		Polar & 0.5 & \textbf{27.7}  & 46.4  & \textbf{28.6}   \\
		Normalized & 0.5 & 27.0  & \textbf{47.9} & 26.9  \\
		\hline
	\end{tabular}
	\caption{Polar centerness vs. normalized centerness}
	\label{tab:pcvsnpc}
\end{table}

\subsection{Comparison to state-of-the-art}
\begin{figure}[t]
	\centering
	\begin{subfigure}{0.49\linewidth}
		\centering
		\includegraphics[width=0.95\linewidth]{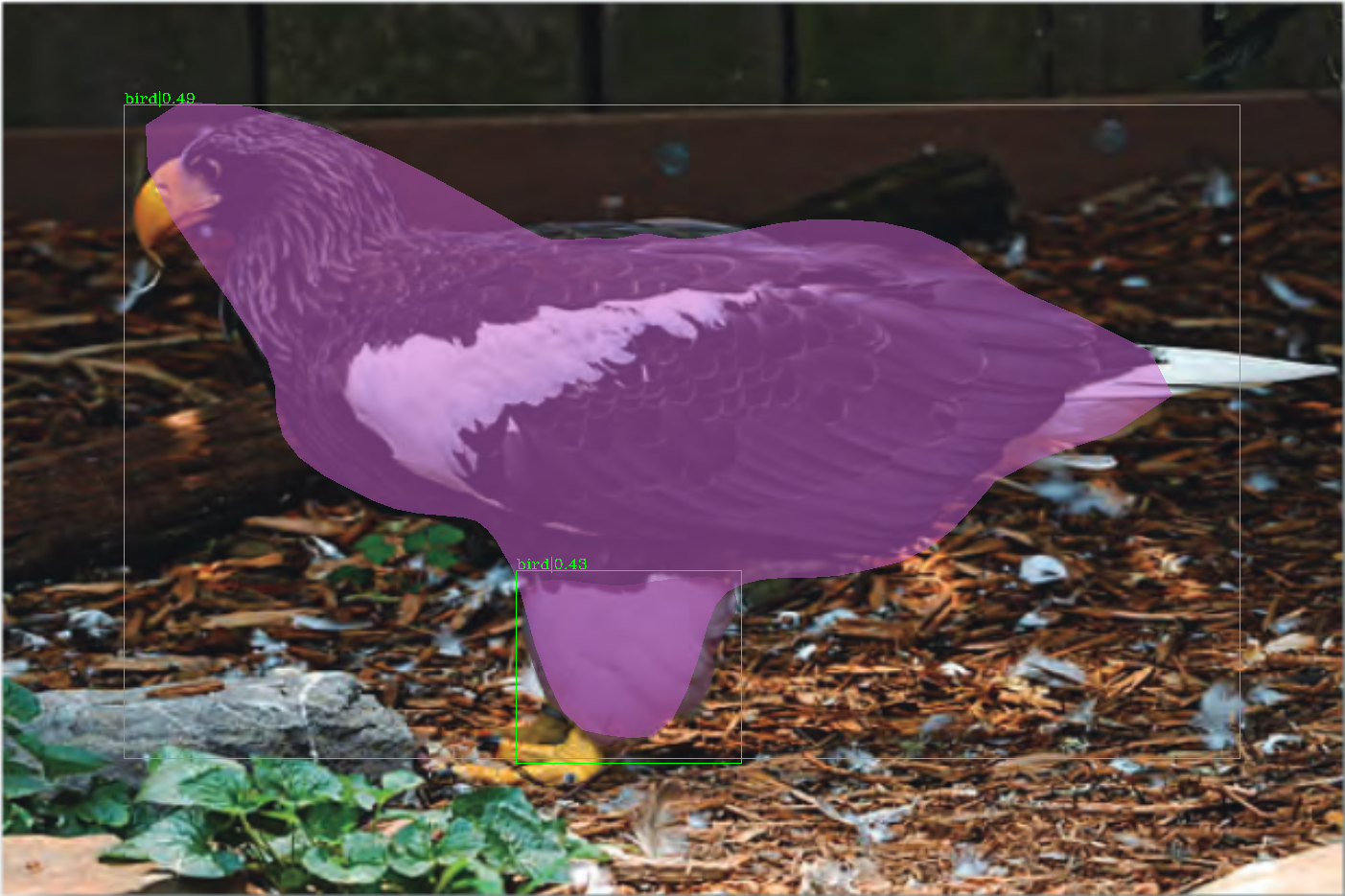}
		\caption{FourierNet-640}
		\label{fig:fouriernet640_1}
	\end{subfigure}
	\begin{subfigure}{0.49\linewidth}
		\centering
		\includegraphics[width=0.95\linewidth]{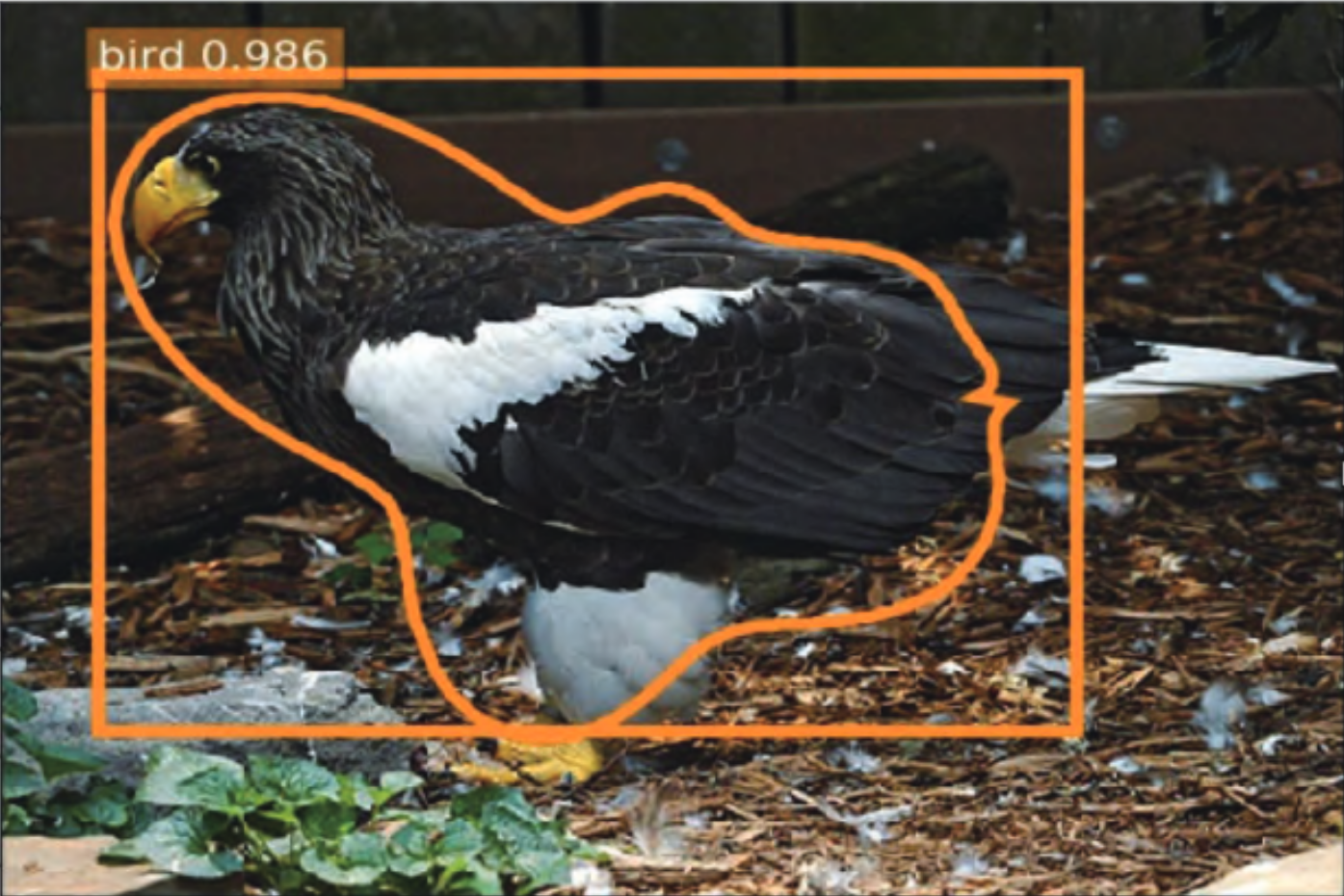}
		\caption{ESE-Seg-416}
		\label{fig:ese_seg_416_1}
	\end{subfigure}
	\begin{subfigure}{0.49\linewidth}
		\centering
		\includegraphics[width=0.95\linewidth]{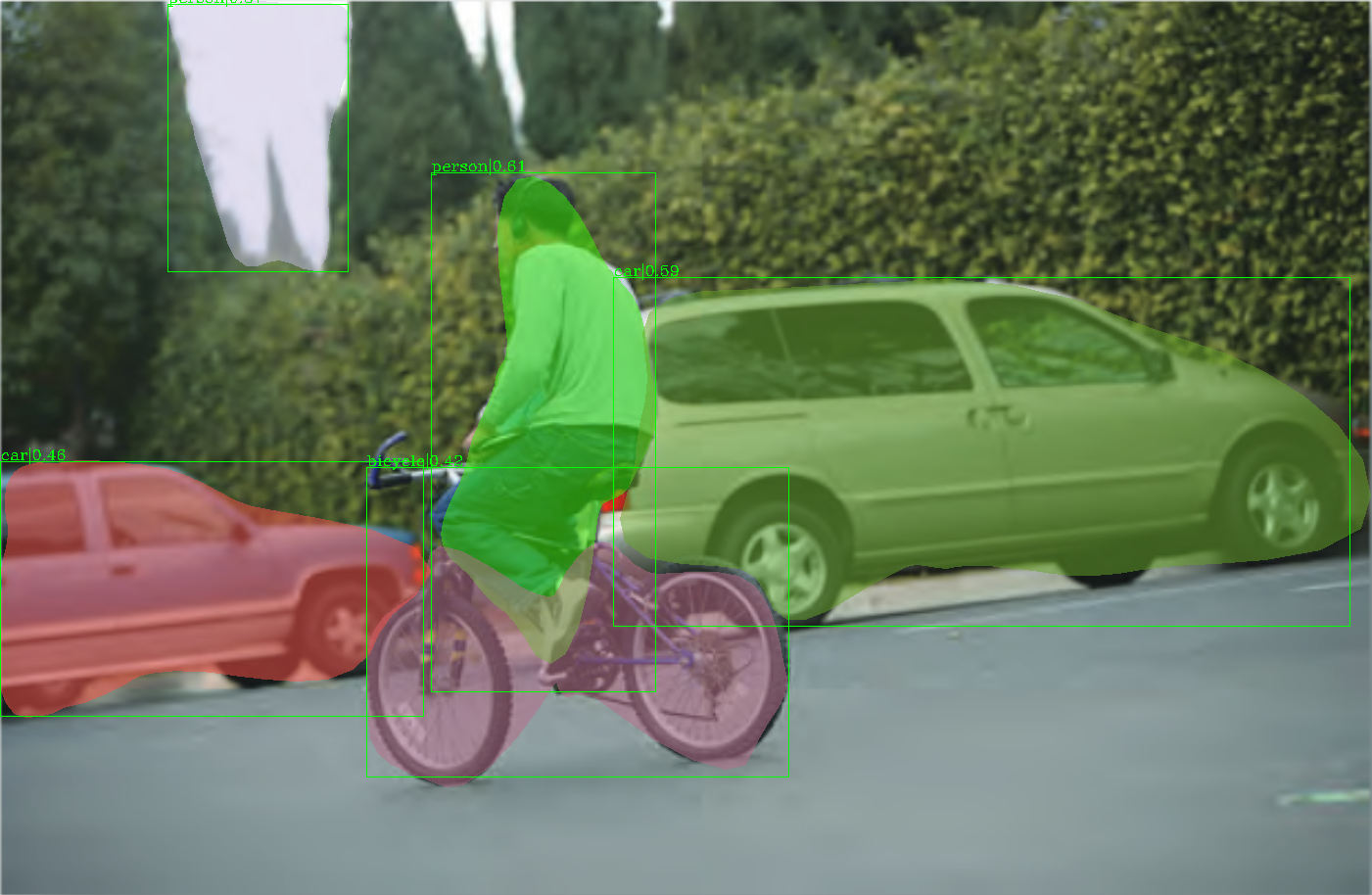}
		\caption{FourierNet-640}
		\label{fig:fouriernet640_2}
	\end{subfigure}
	\begin{subfigure}{0.49\linewidth}
		\centering
		\includegraphics[width=0.95\linewidth]{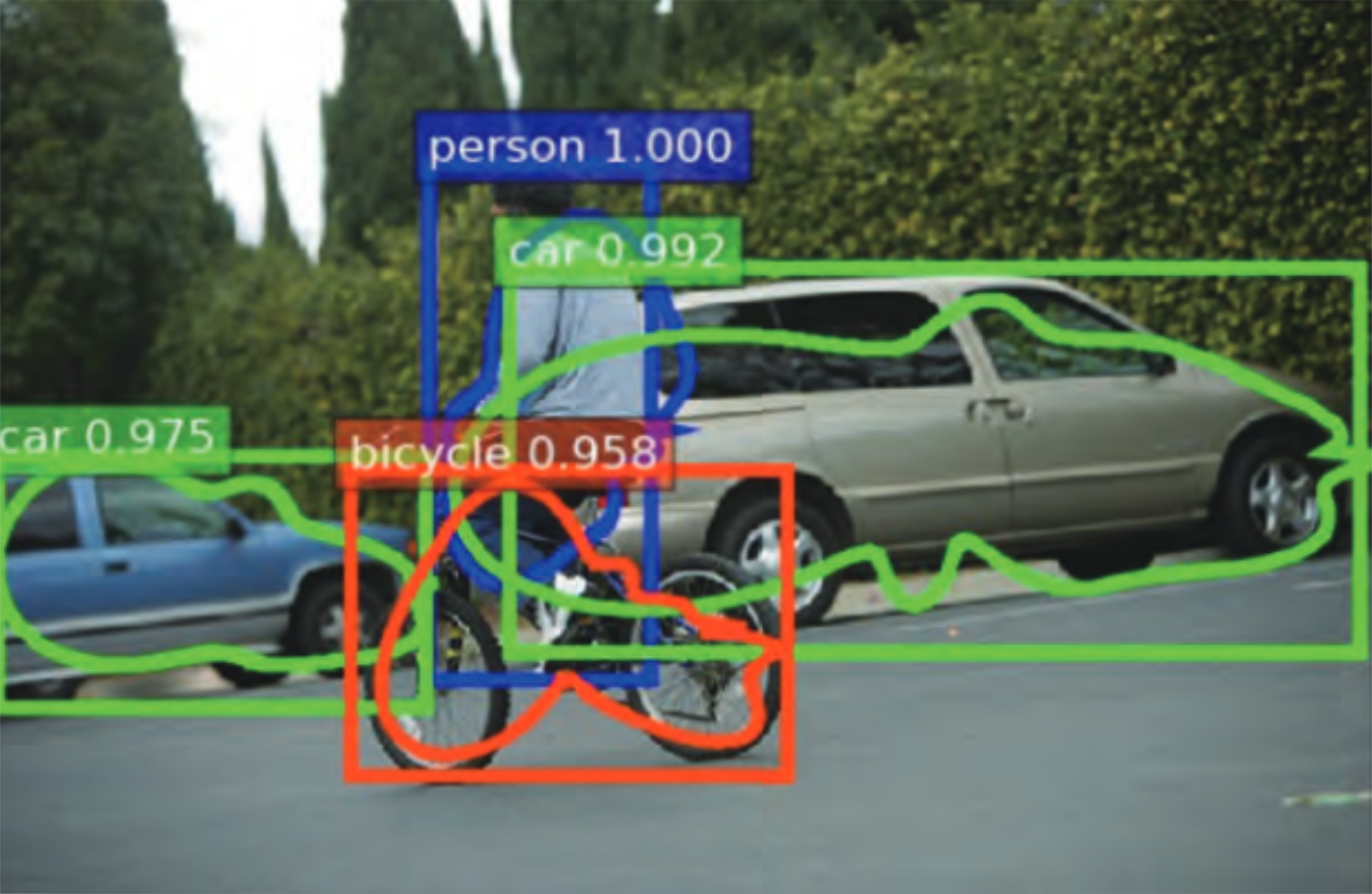}
		\caption{ESE-Seg-416}
		\label{fig:ese_seg_416_2}
	\end{subfigure}
	\caption{A qualitative comparison between FourierNet-640 and ESE-Seg-416.}
	\label{fig:ese_vs_fouriernet}
\end{figure}

\quad We trained a FourierNet-640 with images with a resolution of 640\,x\,360 to compare with ESE-Seg-416, which was trained with a resolution of 416\,x\,416. With a comparable backbone and an equivalent number of parameters, our result is 2.7 mAP higher, and it runs in real-time (figure \ref{fig:ese_vs_fouriernet}).
To compare with the state-of-the-art methods, we trained a FourierNet with a ResNeXt101 backbone \cite{xie2017aggregated}, 90 contour points, and 36 coefficients. The quantitative results are shown in table \ref{tab:sota} and an example of predictions are shown in figure \ref{fig:coeff_compare}.  

Compared to ExtremeNet \cite{zhou2019bottom}, using eight parameters, our results are better, especially with $AP_{L}$ and $AP_{75}$, with an increase of 6.1 and 7.4, respectively. These results show that our mask quality is superior when using few parameters. Moreover, we view that FourierNet is comparable to PolarMask when using the same number of parameters, with a small loss in speed due to the IFFT. Overall, our method is comparable to polygon methods; however, it does not perform as well as binary grid methods.           
\section{Conclusion}
\quad FourierNet is a single-stage anchor-free method for instance segmentation. It uses a novel training technique with IFFT as a differentiable shape decoder. Moreover, since lower frequencies impact the mask the most, we obtained a compact representation of masks using only those low frequencies. Therefore, FourierNet outperformed all methods which use less than 20 parameters quantitatively and qualitatively. Even compared to object detectors, FourierNet can yield better approximations of objects using slightly more parameters. Our FourierNet-640 achieves a real-time speed of 26.6 FPS. We hope this method can inspire the use of differentiable decoders in other applications. 

\bibliographystyle{IEEEbib}
\bibliography{strings,refs}

\end{document}